\documentclass[letterpaper, 10 pt, conference]{ieeeconf}  

\IEEEoverridecommandlockouts                              

\usepackage{graphics} 
\usepackage{epsfig} 
\usepackage{amsmath} 
\usepackage{amssymb}  
\usepackage{graphicx}
\usepackage{slashbox}
\usepackage{subcaption}
\usepackage[table,xcdraw]{xcolor}

\let\labelindent\relax

\usepackage{etoolbox}
\makeatletter
\patchcmd{\@makecaption}
  {\scshape}
  {}
  {}
  {}
\makeatother

\usepackage{caption}
\usepackage{multirow}
\usepackage{pbox}

\usepackage{algorithm}
\usepackage{algorithmic}

\usepackage{enumitem}

\newcommand{\etal}{\textit{et al}. }
\newcommand{\etc}{\textit{etc.} }
\newcommand{\ie}{\textit{i}.\textit{e}. }
\newcommand{\eg}{\textit{e}.\textit{g}. }

\title{\LARGE \bf
Intend-Wait-Cross: Towards Modeling Realistic Pedestrian Crossing Behavior
}

\author{Amir Rasouli$^{*}$ and Iuliia Kotseruba$^{*}$
\thanks{$^{*}$Denotes equal contribution. The authors are with Huawei Noah's Ark Lab, Markham, ON, Canada.
Email: {\tt\small \{amir.rasouli, iuliia.kotseruba\}@huawei.com}}%
}

\begin{document}

\maketitle
\thispagestyle{empty}
\pagestyle{empty}

\begin{abstract}

In this paper, we present a microscopic agent-based pedestrian behavior model Intend-Wait-Cross. The model is comprised of rules representing behaviors of pedestrians as a series of decisions that depend on their individual characteristics (\eg demographics, walking speed, law obedience) and environmental conditions (\eg traffic flow, road structure). The model's main focus is on generating realistic crossing decision-model, which incorporates an improved formulation of time-to-collision (TTC) computation accounting for context, vehicle dynamics, and perceptual noise.

Our model generates a diverse population of agents acting in a highly configurable environment. All model components, including individual characteristics of pedestrians, types of decisions they make, and environmental factors, are motivated by studies on pedestrian traffic behavior. Model parameters are calibrated using a combination of naturalistic driving data and estimates from the literature to maximize the realism of the simulated behaviors. A number of experiments validate various aspects of the model, such as pedestrian crossing patterns, and individual characteristics of pedestrians.

\end{abstract}


\section{Introduction}
Understanding and modeling pedestrian behavior are fundamental for many applications, including traffic flow analysis and control, urban planning, and, more recently, intelligent driving systems. Given the safety critical nature of the latter, there is an acute need for data to train and evaluate proposed algorithmic solutions before they are deployed on the roads. 

One of the most challenging problems in the driving domain is taking into account the behaviors of road users for planning and control. Naturalistic data collection using instrumented vehicles has been an invaluable source for such research \cite{yurtsever2020survey}. Recently, both in academia and industry, simulations emerged as a complementary way of exploring different outcomes of driving scenarios and modeling anomalous events (\eg collisions and near misses) that are difficult to record using traditional methods due to their rarity \cite{fadaie2019state}. However, the utility of the simulations hinges on the behavioral realism of virtual traffic participants. 

Vulnerable categories of road users, such as pedestrians, can potentially benefit the most from safety improvements brought by technology. As road crossing exposes pedestrians to conflicts with other road users, modeling pedestrian crossing decision-making is of top concern. Decades of research indicate that pedestrians exhibit a wide range of behaviors that can be influenced by multiple factors, including pedestrians' individual characteristics (\eg demographics) and environmental conditions \cite{rasouli2019autonomous}. Thus, towards the goal of accurately modeling aspects of pedestrian crossing decision-making, we propose an agent-based simulation motivated by traffic behavior literature.

\begin{figure}
\centering
\includegraphics[width=1\columnwidth]{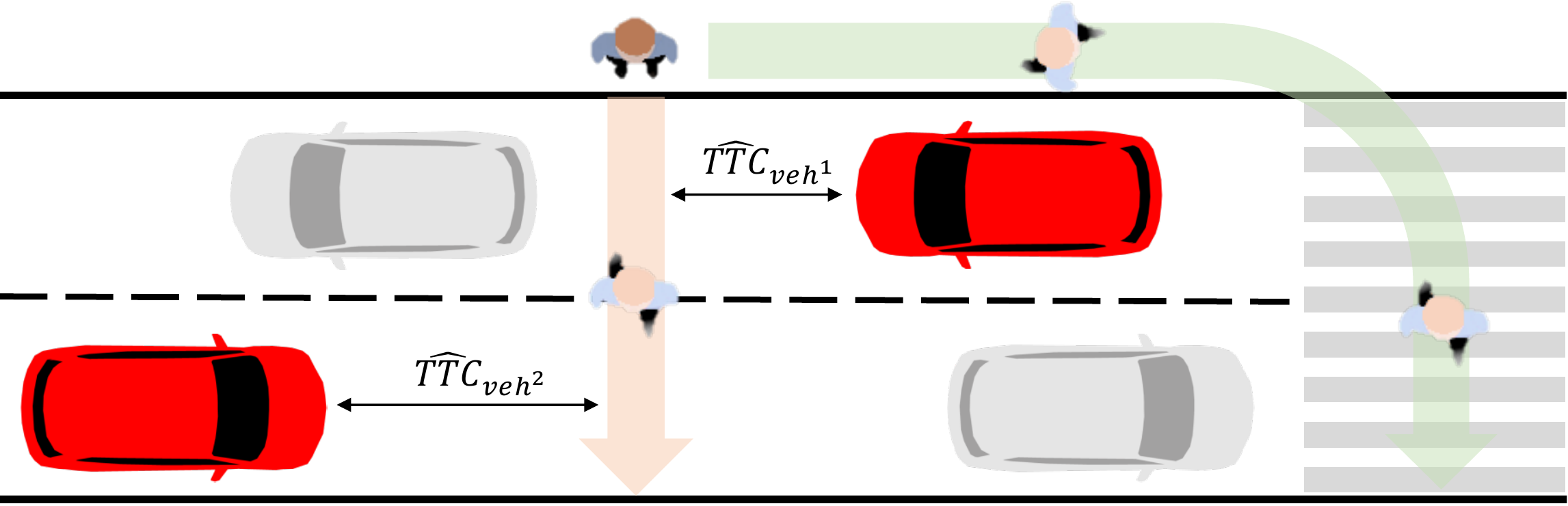}
\caption{Pedestrian decision-making for jaywalking (red arrow) or signalized crossing (green arrow). For jaywalking, time-to-collision (TTC) is estimated for all relevant vehicles (shown in red). Irrelevant vehicles are shown in gray.}
\vspace{-1em}
\label{decoder}
\end{figure}

\section{Related work}

There is a large body of work on modeling interactions between pedestrians and various traffic participants, including other pedestrians and vehicles \cite{camara2020pedestrian}. The simulations are typically subdivided into macro- and microscopic, depending on whether they focus on the movement of traffic or pedestrian flows on the aggregate level or model individual agents \cite{papadimitriou2009critical}. Within microscopic simulations, agent-based models have emerged as effective tools for studying traffic interactions \cite{bazzan2014review}. The agent-based approach is particularly suitable for this purpose since each agent's characteristics, decision-making, and actions are treated individually, and complex interactions between multiple heterogeneous entities give rise to emergent phenomena that may be difficult to program explicitly \cite{bernhardt2007agent}.

A number of models have been proposed for crossing decision-making at the unsignalized crosswalks \cite{lu2016cellular, papadimitriou2016introducing, feliciani2017simulation, zhu2021novel} and during jaywalking \cite{suh2013modeling, papadimitriou2016introducing, wang2021modeling}, scenarios that pose the most risk for pedestrians.

Gap acceptance is one of the most important factors that affect the behaviors of pedestrians and is commonly included in the decision-making models. Usually, the safety of the gap is determined based on the time-to-collision (TTC) and the time it takes the pedestrian to cross the street. Other influences are often included to better capture variations in the observed behaviors of pedestrians. For example, Feliciani \etal \cite{feliciani2017simulation} assign different walking speeds to elderly and adult pedestrians and model crossing as a multi-stage process, where pedestrians assess the safety gap for the nearest lane first and check the safety of the far lane while crossing. Suh \etal \cite{suh2013modeling} distinguish between compliant and gap-seeking pedestrians. The former always cross on a green signal, and the latter may cross during red signal (given the 4-6s gap).

More detailed characteristics of pedestrians are introduced in the model of jaywalking behavior by Wang \etal \cite{wang2021modeling}, where crossing decision depends on the efficiency, safety, and fairness of the route, as well as past crossing decisions. Weights for each property are determined using observational data. Once the decision to jaywalk is made, pedestrians wait for a sufficiently large gap (TTC $>3$s) to start crossing. A different approach to incorporating human factors is proposed by Papadimitrou \etal \cite{papadimitriou2016introducing}. They model crossing as a sequential logit model, where at each step the pedestrian chooses to cross, walk to the intersection, or not to cross. The choices are affected by the road type, traffic density, previous choice, traffic signal, individual's risk-taking propensity, and walking speed (based on the observation that faster pedestrians are more likely to cross mid-block). Parameters of the model are estimated from survey data. 

Although agent-based approaches are effective for simulating pedestrian crossing behavior, they have limited utility for intelligent driving applications. Many aspects of the existing pedestrian models are situation-specific. For instance, environment properties (road layout, traffic flow, \etc) are often rigidly defined. Likewise, pedestrian characteristics are typically limited to a handful of parameters to ensure fit to field data. Therefore, it is unclear whether such simulations can be easily adapted to a wide variety of scenarios and generate a sufficiently diverse population of agents. Other limitations concern crossing behaviors, specifically simplistic estimation of the available gap for crossing in traffic and rigid decision-making steps customized for specific situations.

We address some of these concerns in the proposed model of the pedestrian crossing behavior by introducing the following:
\begin{itemize}
\itemsep0em
\item A highly diverse population of agents with intuitive parameters that can be set based on the field data or using values from the literature; 
\item A discrete choice model for multiple crossing scenarios and non-crossing behaviors; 
\item An improved calculation of time-to-collision that takes into account the relevance of multiple interacting agents, dynamic factors, and pedestrian perceptual noise; 
\item A multi-purpose implementation suitable for modeling a wide range of scenarios within a mature open-source traffic simulation platform.
\end{itemize}

\section{Pedestrian Decision Model}
\begin{figure*}[!t]
\centering
\includegraphics[width=0.95\textwidth]{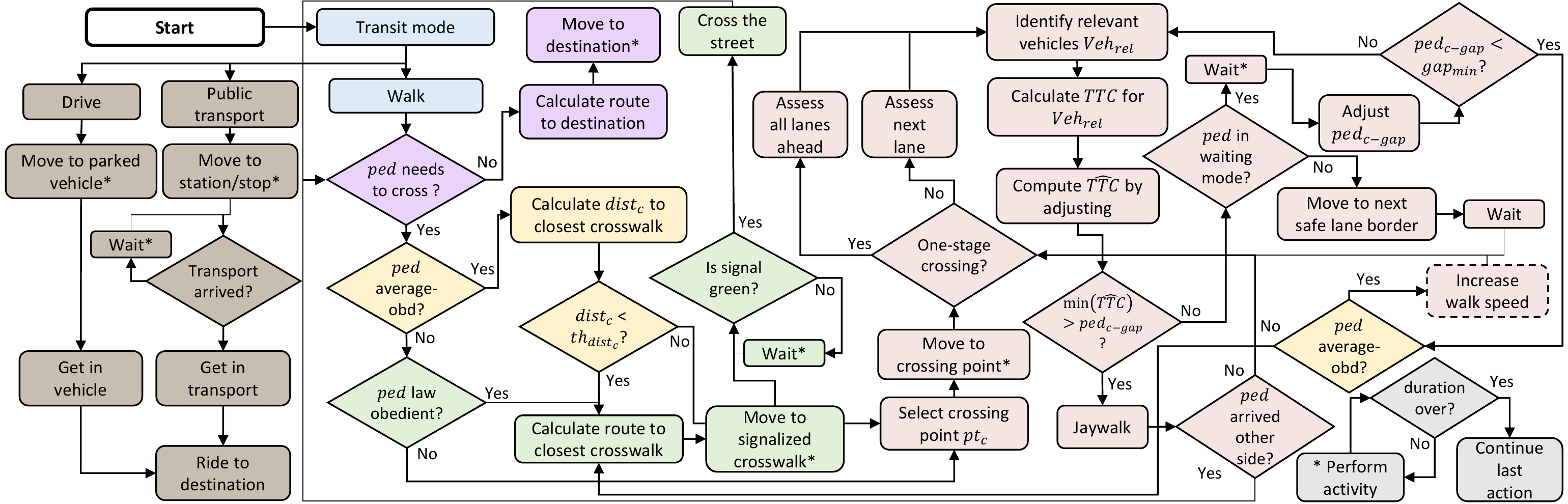}
\caption{The diagram of the proposed pedestrian decision model. Segments of diagram are color-coded as follows: \textit{blue} - agent generation, \textit{brown} - non-walking modes of transit, \textit{gray} - activity while not in transit mode, \textit{purple} - walking with no crossing, \textit{yellow} - crossing for pedestrians with average law obedience, \textit{green} - crossing for law-obedient pedestrians, and \textit{red} - crossing for law-violating pedestrians. Diamond-shaped boxes correspond to yes-no decisions, rectangular boxes indicate actions being performed. Dashed box outline indicates that the action is performed only once.}

\label{fig:decision_model}
\vspace{-1em}
\end{figure*}

Pedestrian decision-making can be categorized by the level into \textit{strategic} (where the pedestrian starts and where they go), \textit{tactical} (decisions that lead to the destination), and \textit{operational} (how the plan is executed) \cite{hoogendoorn2004pedestrian}. In the following sections, we focus on  tactical decision-making, whereas strategic and operational levels are covered in the implementation of the proposed model (see Section \ref{sec:model_calibration}). Figure \ref{fig:decision_model} shows the diagram of the model.

Pedestrians' behaviors are largely determined by their choices, which can be made at a high-level, \eg destination choice, modes of transit, route choice, or low-level, \eg choice of speed, next step, or activity \cite{bierlaire2009pedestrians}. Our model assumes that destination and transit mode are selected randomly, whereas other choices depend on pedestrians' individual characteristics and context. 

\subsubsection{Choice of transit} Four modes of transit are available for pedestrians: \textit{walk}, \textit{drive}, \textit{bus}, and \textit{taxi}. Using walk mode, the pedestrian moves to their destinations entirely on foot, whereas for drive mode, they move to the location of their parked vehicle and drive to their destination. When using public transit including bus and taxi, the pedestrian goes to the designated waiting area, \eg the bus stop, and boards the transport to go to the point closest to their destination. The primary focus of this work is on walking behavior which may involve crossing. 

\subsubsection{Route choice} Route choices are motivated by the pedestrian's characteristics (\eg law obedience, trait), distance to the nearest crosswalk, traffic flow and the destination location (see Section \ref{sec:ped_char}).

\subsubsection{Activity choice} At any point during waiting or walking, pedestrians may engage in an \textit{activity}, \eg smoking or taking a photo, for an arbitrary duration. Performing such activities does not impact pedestrians' decisions to cross.

\subsubsection{Choice of speed} Pedestrian average speed is determined by their type and trait (as described in Section \ref{sec:ped_char}), \eg the more aggressive a pedestrian is, the faster they would walk/cross. For each pedestrian, walking speed is randomly sampled from a skewed normal distribution, 
$$ ped_s \sim 2 \times \mathcal{N}(\mu_{ps},\sigma_{ps}^{2}) \times \Phi (\alpha \mu_{ps}, \sigma_{ps}^{2})$$

\noindent where $\mu_{ps}$ and $\sigma_{ps}$ are mean and standard deviation of the speed for a given pedestrian type, $\alpha$ is skew factor, and $\Phi$ is CDF of the normal distribution. The skew factor is set to the mean value, with the negative or positive sign for conservative and aggressive pedestrians, respectively, and zero for average pedestrians. Pedestrian speed may also temporarily increase during crossing depending on how long they have been waiting at the curb (see Section \ref{sec:jaywalk}). 

\subsubsection{Choice of next step} is determined dynamically depending on the route choice and scene dynamics, particularly when the pedestrian intends to jaywalk (see Section \ref{sec:jaywalk}).
	
\section{Pedestrian characteristics}
\label{sec:ped_char}
Personal characteristics (\eg walking speed, road assessment, gap acceptance) impact pedestrians' crossing behaviors \cite{rasouli2019autonomous}. In our model, we consider six main characteristics, namely pedestrian \textit{type}, \textit{trait}, \textit{law obedience}, \textit{gap acceptance}, \textit{crossing pattern}, and \textit{perceptual noise}.
\subsubsection{Type} Pedestrian type $ped_{type}$ captures the age group and mobility of pedestrians. There are three pedestrian types in our model: adult, child, and elderly. Different types of pedestrians have different walking speeds affecting their gap acceptance and crossing.
\subsubsection{Trait} Each pedestrian can have one of the three pre-defined traits $ped_{trait}$: \textit{aggressive}, \textit{conservative}, and \textit{average}. The trait of a pedestrian affects their walking speed and gap acceptance and, consequently, their crossing decision-making. For example, aggressive pedestrians walk faster and accept shorter gaps to cross, therefore crossing earlier than conservative ones.
\subsubsection{Law obedience} Pedestrian law obedience $ped_{lo}$ determines their route choice. \textit{Law-obedient} pedestrians always cross at designated crosswalks and obey traffic signals, whereas \textit{law-violating} ones always jaywalk. An \textit{average} person may be law obedient or violating depending on the context. In our model, an average person is obedient if their distance to the closest designated crosswalk, $dist_{c}$, is less than a pre-defined threshold, $th_{dist_{c}}$.
\subsubsection{Gap acceptance} Pedestrians accept different time gaps for crossing. This is relevant for pedestrians when they intend to jaywalk. Similar to walking speed, the initial gap acceptance of the pedestrian $ped_{gap}$ is sampled from a skewed distribution with a mean and standard deviation of the defined $gap_{range}$ and skew factor set to a positive or negative mean gap for conservative and aggressive traits, respectively.
\subsubsection{Crossing pattern} 
\begin{figure}[!t]
\centering
\includegraphics[width=1\columnwidth]{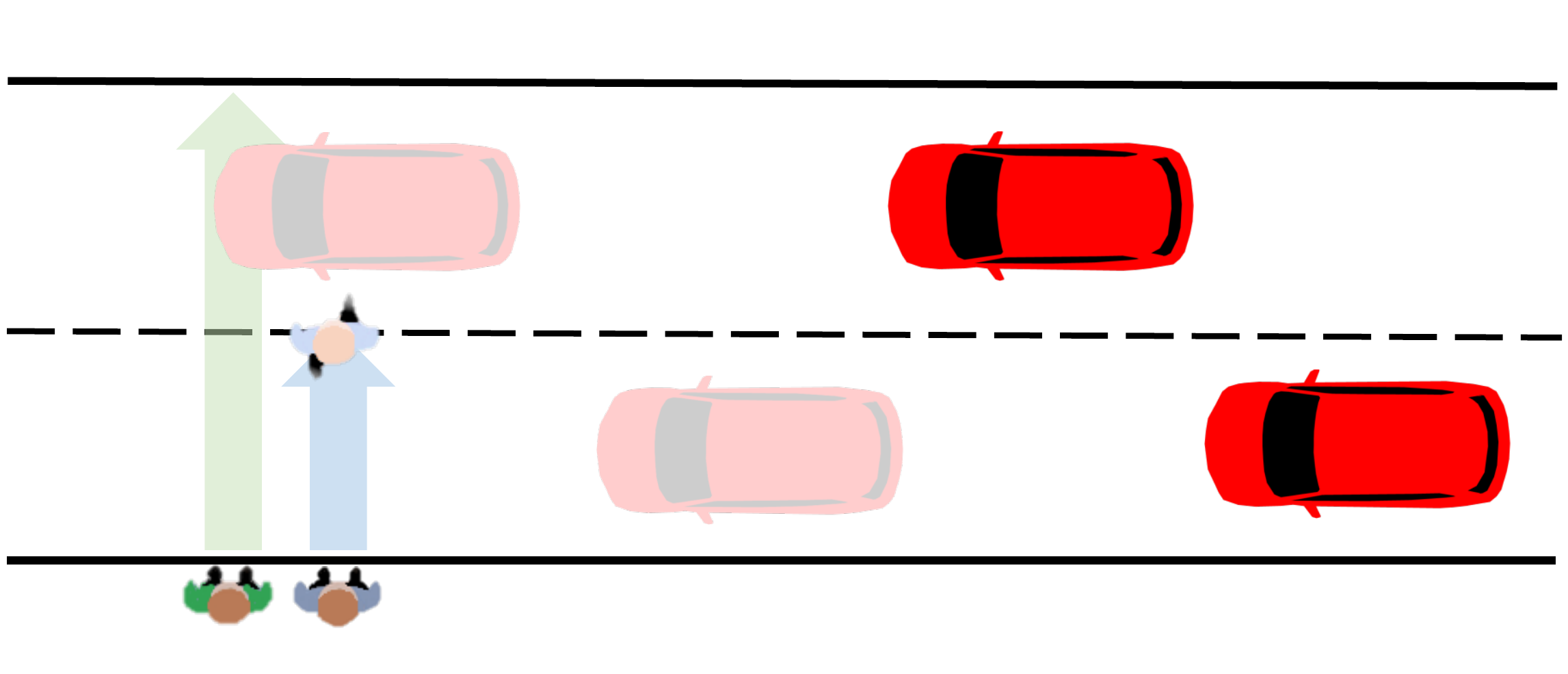}
\caption{Illustration of different \textit{crossing patterns}. Future locations of the vehicles are shown with transparent vehicle outlines. Pedestrian performing a \textit{one-stage} crossing (green arrow) cannot proceed because the vehicle in the far lane intersects its path. Pedestrian with a \textit{rolling gap} strategy (blue arrow) begins to cross since the vehicle in the nearest lane is sufficiently far.}
\label{fig:crossing_pattern}
\vspace{-1em}
\end{figure}

This characteristic represents crossing preference of the pedestrian as one of two categories: \textit{one-stage}, when the pedestrian waits until all lanes are clear to cross, and \textit{rolling gap} if the pedestrian begins to cross as soon as the gap at the nearest lane becomes available \cite{brewer2006exploration} (see Figure \ref{fig:crossing_pattern}). Pedestrians also have a parameter that determines whether they wait for stopped vehicles that block their way or move around them in order to cross.
\subsubsection{Perceptual noise} Pedestrians differ in terms of their ability to assess their surroundings, \ie  how accurately they can estimate the distance and speed of other agents in the environment. We sample a perception error rate from a normal distribution, $ped_{p-noise} \sim \mathcal{N}(0,1)$.  This factor ultimately impacts the gap pedestrians accept to cross the road (see Section \ref{sec:noise_ttc}). 

\section{Pedestrian walking behavior}

Walking behavior is at the core of the proposed model. If the selected route does not involve crossing, a pedestrian simply follows the shortest path to the destination. Otherwise, they may need to cross one or more times depending on the road layout and location of the final destination. Below we provide a detailed overview of the decision-making during various types of crossing.

\subsection{Crossing at designated crosswalks} There are two possible options: a) crossing only at the signalized crosswalks and b) crossing at any designated crosswalks. If the first option is set, law-obeying pedestrians prioritize signalized crossing, and only if none are available along the route will cross at any designated crosswalk. In either case, pedestrians choose crosswalks that are closest to their final destinations. Once the crosswalk is selected, the shortest path from the current location of the pedestrian to the crosswalk is calculated. Next, the pedestrian moves to the crosswalk and crosses the street following a traffic signal or if vehicles are yielding at the unsignalized crosswalk.
\label{sec:jaywalk}
\subsection{Jaywalking} 
A pedestrian that jaywalks starts with selecting a crossing point. In our model, the pedestrian first randomly selects a crossing point $pt_c$  close to the destination. Then, the shortest route to the crossing point is computed.

\subsubsection{Intend to cross} Once the pedestrian arrives at the crossing point, their status changes to intending to cross. Then, the pedestrian moves to the curbside and waits to cross. 

\subsubsection{Wait}
\label{sec:wait}
 During this phase, the pedestrian assesses the traffic and waits for a suitable crossing gap $ped_{c-gap}$. Depending on the \textit{crossing pattern}, the pedestrian will either wait for all lanes ahead to have a sufficient gap (one-stage) or the nearest lane only (rolling gap) (see Figure \ref{fig:crossing_pattern}).

During the wait phase, $ped_{c-gap}$ is dynamically updated depending on the pedestrian's wait time $ped_{wt}$ prior to crossing, 
$$ped_{c-gap} = \max (gap_{min}, f_{wt}(ped_{gap},ped_{wt})) $$

\noindent where $gap_{min}$ is the minimum accepted gap threshold and $f_{wt}$ is a function of pedestrian gap acceptance and wait time. Here, the initial gap acceptance of pedestrian is reduced linearly by a constant value $wt_{const}$ for each second while the pedestrian is waiting. 

\begin{figure}[!t]
\centering
\includegraphics[width=1\columnwidth]{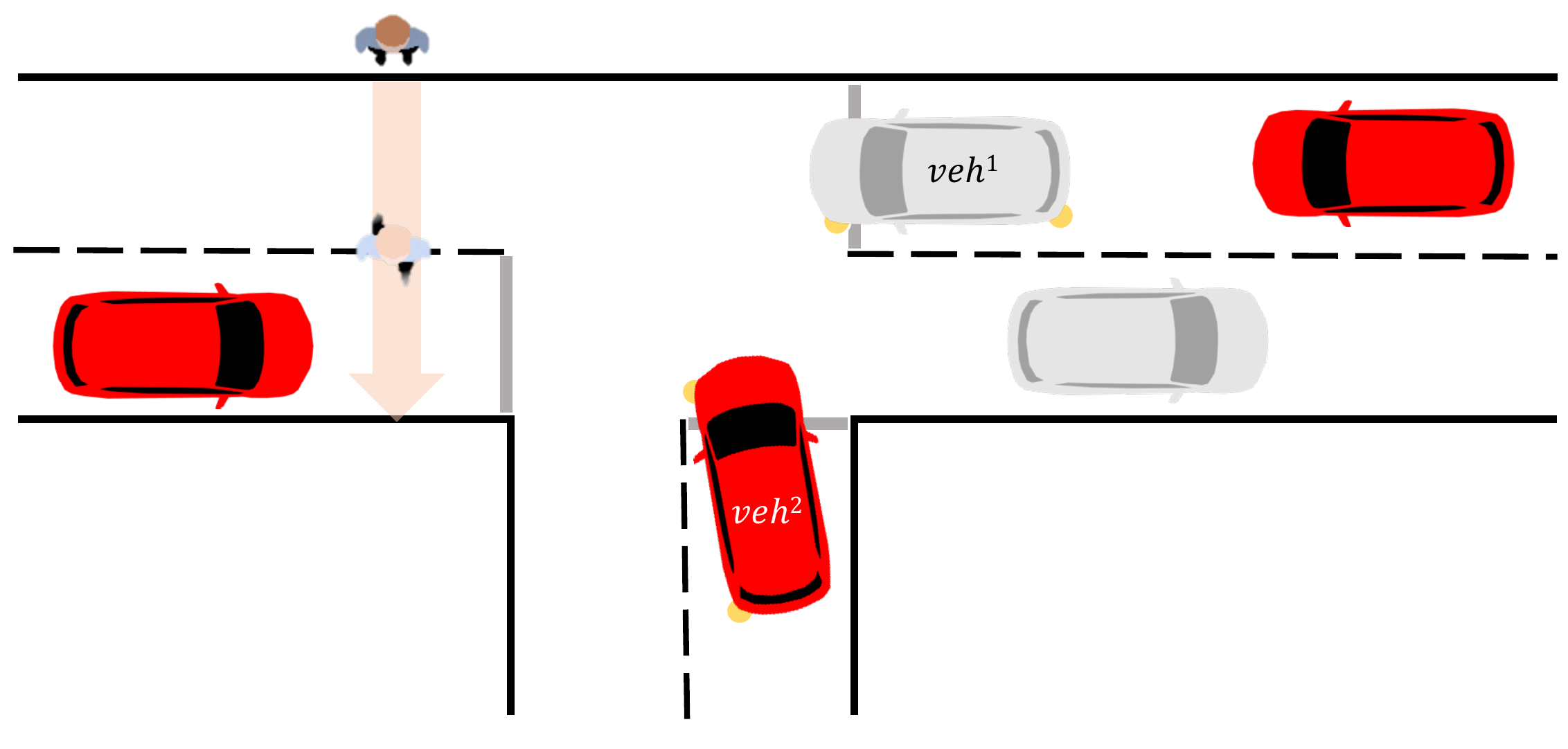}
\caption{Illustration of vehicle relevance for TTC computation. Red vehicles are relevant and gray ones are irrelevant for the jaywalking pedestrian. $veh^{1}$ has left turn signal on showing the intention of turning left, therefore it is irrelevant as opposed to the vehicle behind it that intend to go straight. In the case of $veh^{2}$ since the vehicle is signaling and turning towards the path of the pedestrian is relevant.}
\label{fig:ttc_relevance}
\vspace{-1em}
\end{figure}

\subsubsection{Compute time-to-collision (TTC)}

\begin{figure}[!t]
\centering
\includegraphics[width=1\columnwidth]{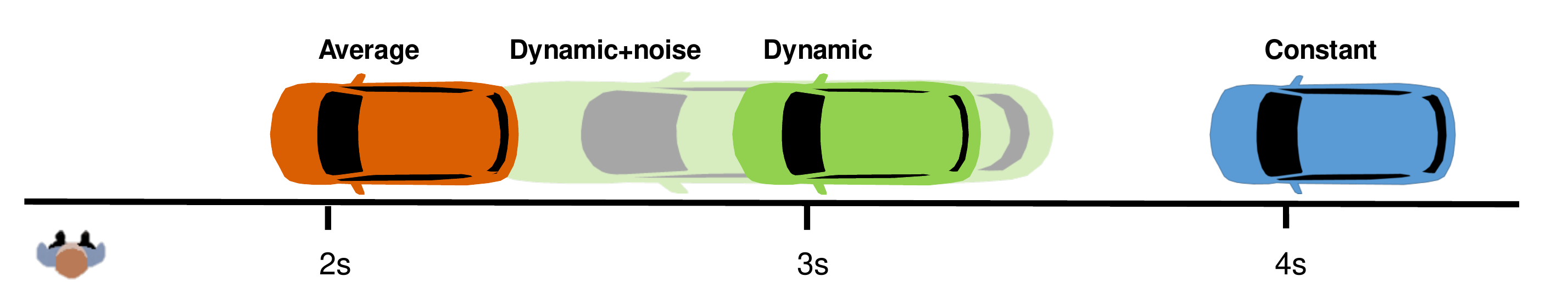}
\caption{Comparison of different TTC computation methods for a given scenario. Vehicle is located $15m$ away from the pedestrian moving at $4m/s$ and accelerating at $1m/s^2$.  }
\label{fig:ttc_mode}
\vspace{-1em}
\end{figure}

To compute the crossing gap, the minimum TTC of the approaching vehicles should be calculated first. If $ped_{c-gap} < minTTC$, the pedestrian starts crossing. TTC calculation involves the following steps:

\textbf{Identifying relevant vehicles}. The first step is to identify relevant vehicles, \ie those that pose a potential threat to the pedestrian. Intuitively, these are the vehicles on routes that potentially intersect with the pedestrian's path (positional relevance) and those that intend to take intersecting routes with pedestrians (intentional relevance). For a two-way street, positionally relevant vehicles $\operatorname{Veh_{rel}}$ are expressed by,
\begin{gather*}
  \operatorname{Veh_{rel}} = \forall veh^i | -\pi < \theta_p, \theta_{v^i} < \pi \quad \wedge \notag\\
   sign(\theta_p) \neq sign(\theta_{v^i}),  i \in {1,2,...,n} \notag\\
		\theta_p = \measuredangle p{v^i} - \phi_{ped} \notag\\
		\theta_{v^i} = \measuredangle {v^i}p - \phi_{veh^i}
\end{gather*}

\noindent where $\phi_{ped}$ and $\phi_{veh^i}$ are the pedestrian and vehicle orientations respectively.  $\measuredangle p{v^i}$ and $\measuredangle {v^i}p$ represent the angles between the pedestrian and vehicle and vice versa given by, 
$$ \measuredangle rt = arctan(t_x - r_x, t_y - r_y)\times 180 / \pi.$$

Once the positional relevance of the vehicles is determined, their signal status is considered to further refine the list of relevant vehicles according to their intentional relevance. For example, if a vehicle is moving towards the pedestrian from the other side of an intersection, it is only relevant if it is moving straight. However, if the vehicle has its turn signal on, it is no longer relevant (see Figure \ref{fig:ttc_relevance}).

\textbf{Calculating TTCs for relevant vehicles.} The common approach to computing TTC is by dividing the distance between the front of the approaching vehicle and the pedestrian by the vehicle's current speed (constant) \cite{chao2015vehicle,zheng2015modeling,schroeder2011event}. Alternatively, the average of current speed and the maximum road speed limit (average) can be used to account for changes in speed. However, these approaches provide only a crude estimate. First, they do not take into account accurate vehicle acceleration, \eg if the vehicle is starting to move from a stationary position or to slow down. Second, considering the distance only to the front of the vehicle is insufficient since the pedestrian can also hit a side of the vehicle, especially a longer one, such as a bus or a truck. 

For the reasons above, we define a dynamic TTC model (see Figure \ref{fig:ttc_mode} for a comparison).  We start by measuring the time it takes the vehicle to achieve its maximum speed, 
$$  t_{max_s} = (max_s - veh_s)/veh_{accl}$$

\noindent where  $veh_{accl}$ and $veh_s$  are vehicle acceleration and its current speed. $max_s$ is given by 
$$ max_s = min(road_{max_s}, veh_{max_s})$$

\noindent where $road_{max_s}$ is the given road speed limit and $veh_{max_s}$ is the maximum speed of the vehicle. As highlighted here, we assume the vehicles do not exceed the speed limit.

Next, we compute the time it takes the vehicle to arrive to the pedestrian as,
$$ t_{v\rightarrow p} = (-veh_s + \sqrt{veh_s^2 - 2\times veh_{accl}\times veh_{dist}})/ veh_{accl}$$

\noindent $veh_{dist}$ is the distance of the vehicle to the point where its and the pedestrian's paths intersect.  If $t_{v_{max_s}} > t_{v\rightarrow p}$, then $t_{v\rightarrow p}$ is considered as TTC of the given vehicle. Otherwise, 
$$ TTC =  t_{max_s} + t_{const_s} $$
$$ t_{const_s} = (veh_{dist} - dist_{max_s})/ max_s$$

\noindent where $dist_{max_s}$ is the distance that takes the vehicle to reach the maximum speed given by,
$$dist_{max_s} = 0.5 \times (veh_s + max_s) \times t_{max_s}.$$

\textbf{Adjusting TTCs.} The TTC of each relevant vehicle is computed per lane based on the distance between the front of the vehicle and the projected location of the pedestrian on the given lane. The TTCs, however, should be adjusted to reflect both the time it takes the pedestrian to arrive at the given lane and the length of the vehicle, \ie the time it takes the back of the vehicle to pass a given point. To achieve this, we first measure the travel time of the pedestrian to the given lane and subtract it from the TTCs of vehicles on that lane resulting in $\hat{TTC}$. Here, if $\hat{TTC}_{veh} >= 0$, then the vehicle does not pass the pedestrian by the time they arrive at the given lane. Such vehicles remain relevant. 

When $\hat{TTC}_{veh} < 0$, only the front of the vehicle passes the pedestrian. We further measure $t_{veh-rear}$, the time it takes for the rear bumper of the vehicle to pass. If $ \hat{TTC} + t_{veh-rear} < 0$, then the vehicle completely passes the pedestrian and becomes irrelevant (\ie $\hat{TTC}_{veh} = \infty$). Otherwise, pedestrian will collide with the side of the vehicle, so we set $\hat{TTC}_{veh} = 0$.

Note that in the case of the vehicles that are stopping and blocking the pedestrian path, their TTC values are excluded if the pedestrian's crossing pattern induces them to move around the stopped vehicles.

\subsubsection{Waiting to cross for too long}
\label{sec:wait_too_long}
If the pedestrian waits too long before crossing, \ie $f_{wt}(ped_{gap},ped_{wt})<gap_{min}$, they change their crossing behavior according to their law obedience. Law-violating pedestrians will increase their crossing speed $ped_{crs}$ by $sp_f$ times (mimicking running across the street), allowing them to accept shorter gaps. On the other hand, average pedestrians will switch their status to law-obedient, find the nearest designated crosswalk, and alter their route accordingly. 

\subsubsection{Jaywalking} A pedestrian will begin to jaywalk when crossing gap is smaller than the minimum TTC across all relevant vehicles, \ie $ped_{c-gap} < min(\hat{TTC}_{Veh_{rel}})$. It should be noted that the relevance of the vehicle might change during jaywalking. For example, a vehicle might turn into the road where the pedestrian is crossing or, perhaps, a stationary vehicle starts moving. This means that pedestrians must continuously estimate $min(\hat{TTC}_{Veh_{rel}})$ and only proceed to cross when it is safe to do so. 

Depending on their \textit{crossing pattern}, pedestrians will wait either for all lanes to be clear (one-stage) or just the lane nearest to them (rolling gap). In both cases, pedestrians re-calculate the minimum TTC after taking each step towards the other side of the street. If it is safe to move forward, they do so in the next step, otherwise, they move to the edge of the closest safe lane (next or previous) and wait for a gap. Waiting in the middle of the road can happen even if the pedestrian performs one-stage crossing due to the dynamic nature of the traffic and noisy estimates of the TTC. Note that neither pedestrian gap acceptance nor crossing behavior do not change during this phase (as was the case during the initial wait phase, see Section \ref{sec:jaywalk}). Therefore, $ped_{c-gap}$ and the final walking speed determined during the wait phase are used. 

\subsubsection{Noise in TTC computation}
\label{sec:noise_ttc}
In reality, pedestrians do not have perfect vehicle speed and distance estimates. Therefore their perception of vehicles is noisy, often resulting in underestimated time-to-collision. To account for perceptual noise in the calculation of TTCs, we follow the model proposed in \cite{caird1994perception}.

For TTC calculation, we compute  mean and standard deviation of the perceived (or judged) TTC values as follows,
$$ pTTC_\mu = 0.7 + 0.56\times TTC$$
$$ pTTC_\sigma = 0.17 \times TTC + 0.49.$$

Next, using reparameterization technique, we compute the perceived TTC values for a given pedestrian,
\vspace{-0.25em}

$$ ped_{pTTC} = pTTC_{\mu} + ped_{p-noise}\times pTTC_\sigma.$$
For more realistic modeling, in cases where the vehicle is in close proximity of the pedestrian, we consider a cut-off TTC noise threshold $TTC_{noise-th}$ below which the noise impact is reduced to zero.

\section{Evaluation}
\subsection{Implementation}
\begin{figure}[!t]
\centering
\includegraphics[width=1\columnwidth]{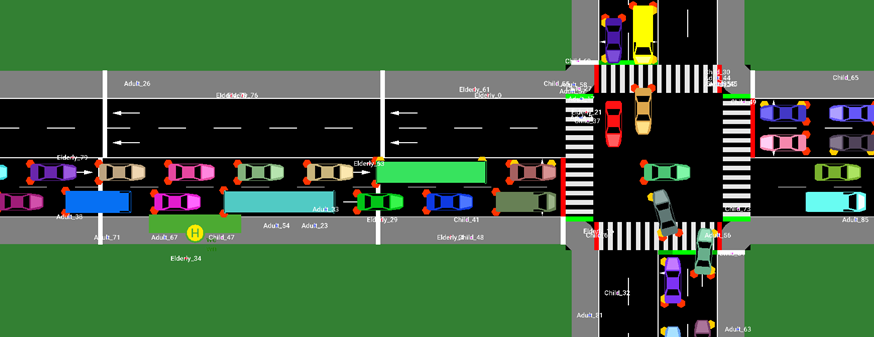}
\caption{A sample of simulation environment with different agents. Pedestrians can be seen by their ids.}
\vspace{-1em}
\label{fig:sim_env}
\end{figure}

For evaluation, we used open-source traffic simulation platform SUMO (v1.9.0) \cite{lopez2018microscopic}. Some of the controls were handled by SUMO's default behavior, including traffic light phase changes, vehicle movements, interactions between pedestrians, and pedestrian movements with the exception of jaywalking action which was manually controlled by the proposed model.\\
\textbf{Road setup}. In the road network, the distance between all nodes was set to $30m$. Each road direction has two lanes, each $3.5m$ wide. Both sides of the street have $3m$ sidewalks, and the speed limit of the street was set to $50km/h$ (or $13.89m/s$), typical for urban areas. 

A default 4-way intersection has 2 road segments, intersecting at a straight angle. Horizontal segment has 9 nodes, and vertical one has 7 nodes. 5 signalized crosswalks are placed: one at the center of the intersection, and one at each connecting road.\\
\textbf{Agents}. We implemented three pedestrians types, adults (A), children (C), and elderly (E), which have different walking speeds and dimensions (children are smaller than the other two). For vehicles, we used the following default vehicle types from SUMO: passenger (P), bicycle (Bi), truck (T), bus (Bu), and motorcycle (M).\\
\textbf{Routing}. All routes were computed within our model. For pedestrians, departure and destination points were randomly selected from available network edges, and the shortest path between the two was calculated subject to pedestrian law-obedience status. In the case of vehicles, departure and destination points were selected from the network endpoints, and the departure lanes were randomized when a new vehicle was generated. A sample view of the simulation environment with different pedestrian and vehicle types and available infrastructure elements (bus stops, crosswalks, traffic lights) is shown in Figure \ref{fig:sim_env}.

\subsection{Model Calibration}
\label{sec:model_calibration}
The default simulation and model parameters used in our experiments are listed in Table \ref{tab:default_params}. In order to produce more realistic scenarios, we set the vehicle and pedestrian types according to the statistics extracted from the naturalistic driving dataset PIE \cite{rasouli2019pie}. Mean pedestrian walking speeds for different pedestrian types are set based on the field study by Knoblauch \etal \cite{knoblauch1996field}. Since pedestrian gap acceptance can differ widely depending on the country and environment \cite{rasouli2019autonomous}, we set the default gap acceptance range following the study conducted by Schmidt \etal \cite{schmidt2009pedestrians} in Europe, which is in agreement with an earlier European study by Ashworth \cite{ashworth1970analysis} and North American one by Brewer \etal \cite{brewer2006exploration}.  

Intrinsic pedestrian parameters, such as trait, law-obedience, and crosswalk distance threshold, were difficult to estimate from the available naturalistic data. Therefore, the remaining parameters were set empirically to achieve a balanced set of crossing behaviors. 

\begin{table}[]
\centering
\caption{Default simulation and decision model parameters}
\label{tab:default_params}
\resizebox{\columnwidth}{!}{%
\begin{tabular}{lll}
\rowcolor[HTML]{FFFFFF} 
Parameters          & Value(s)                  & Description               \\ \hline
\rowcolor[HTML]{FFFFFF} 
Simulation          &                           &                           \\ \hline
\rowcolor[HTML]{EFEFEF} 
$ped_{trait}$ \%    & Con(40), Agg(40),Avg(2)   & \% of ped traits          \\
\rowcolor[HTML]{FFFFFF} 
$ped_{lo}$ \%       & LV(40), LO(40),Avg(20)    & \% of ped law-obedience   \\
\rowcolor[HTML]{EFEFEF} 
$ped_{type}$ \%     & A(89), C(1), E(10)        & \% of ped types           \\
\rowcolor[HTML]{FFFFFF} 
$\mu_{ps}$ m/s      & A(1.51), C(1.48), E(1.25) & mean walking speeds       \\
\rowcolor[HTML]{EFEFEF} 
$\sigma_{ps}$ m/s   & 0.14                      & std dev  of walking speed \\
\rowcolor[HTML]{FFFFFF} 
Veh \% &
  \begin{tabular}[c]{@{}l@{}}P(30), Bu(20), Bi(30), \\ T(11), M(1)\end{tabular} &
  \% of vehicle types \\
\rowcolor[HTML]{EFEFEF} 
$veh_{max_s}$ m/s   & Default SUMO              & max vehicle speed         \\
\rowcolor[HTML]{FFFFFF} 
$road_{max_s}$  m/s & 13.89                     & road speed limit          \\
\rowcolor[HTML]{EFEFEF} 
Sim step s          & 0.1                       & simulation step length    \\ \hline
\rowcolor[HTML]{FFFFFF} 
Decision-making     &                           &                           \\ \hline
\rowcolor[HTML]{EFEFEF} 
$gap_{range}$ s     & {[}3, 8{]}                & gap acceptance range      \\
\rowcolor[HTML]{FFFFFF} 
$gap_{min}$   s     & 2                         & min accepted gap          \\
\rowcolor[HTML]{EFEFEF} 
$wt_{const}$  s     & 1                         & wait time constant        \\
\rowcolor[HTML]{FFFFFF} 
$dist_{c_{range}}$ m &
  {[}30, 60{]} &
  \begin{tabular}[c]{@{}l@{}}range of distance threshold \\ to crosswalks for avg peds\end{tabular} \\
\rowcolor[HTML]{EFEFEF} 
$sp_f$ &
  3 &
  \begin{tabular}[c]{@{}l@{}}speed up factor for \\ violating pedestrians\end{tabular} \\
\rowcolor[HTML]{FFFFFF} 
Crossing \%    & 50                        & \% crossing pedestrians   \\
\rowcolor[HTML]{EFEFEF} 
TTC method            & Dynamic                   & default TTC method   \\
\rowcolor[HTML]{FFFFFF} 
$TTC_{noise-th}$  s    & 0.3                        & TTC noise cut-off threshold  
      
\end{tabular}%
}
\vspace{-1em}
\end{table}

\subsection{Effects of law obedience and traits}

\begin{figure}[!htbp]
\centering
\begin{subfigure}[t]{0.45\columnwidth}
\centering
\includegraphics[height=1.5in]{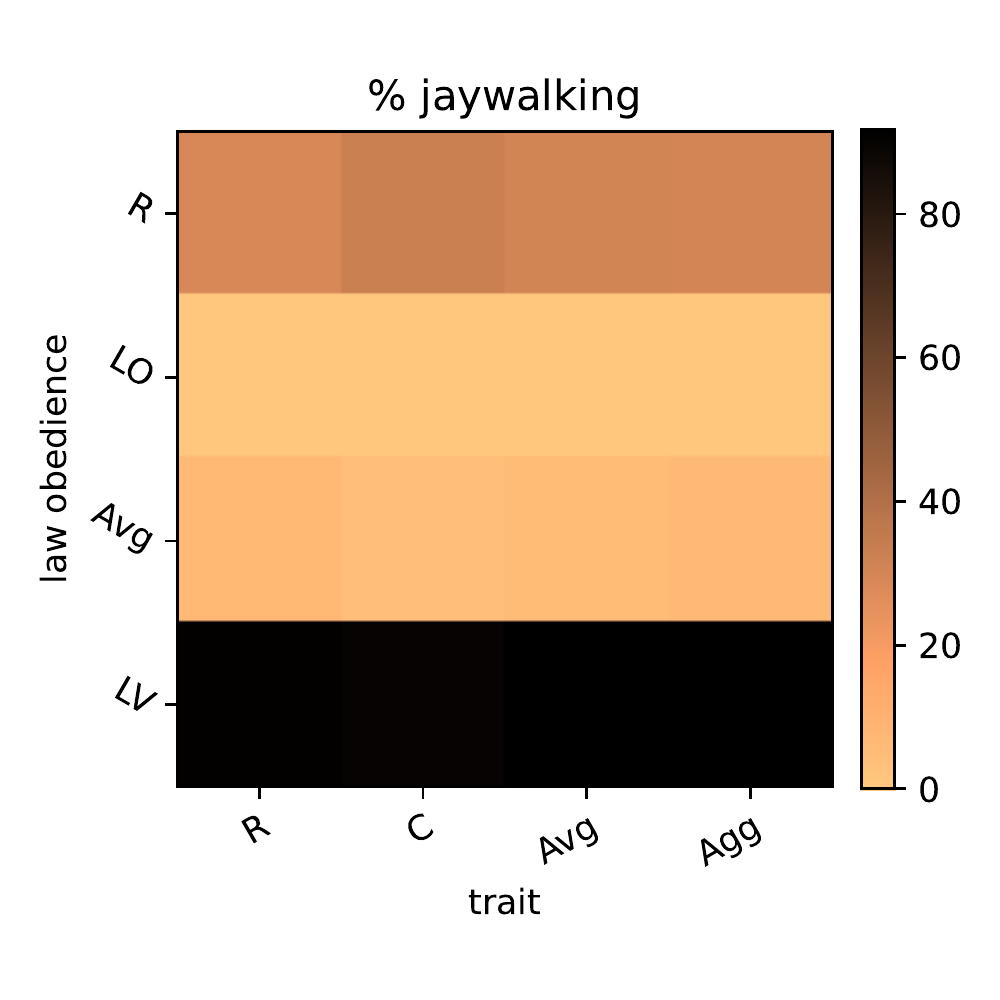}
\caption{}
\label{fig:lo_vs_trait}
\end{subfigure}
~
\begin{subfigure}[t]{0.45\columnwidth}
\centering
\includegraphics[height=1.5in]{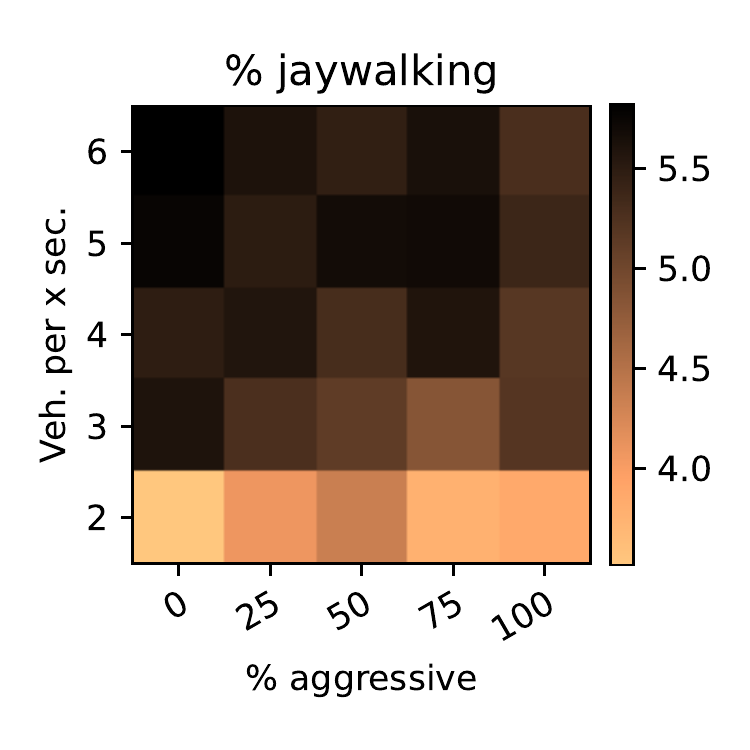}
\caption{}
\label{fig:average_vs_aggressive}
\end{subfigure}

\begin{subfigure}[t]{1\columnwidth}
\centering
\includegraphics[height=1.7in]{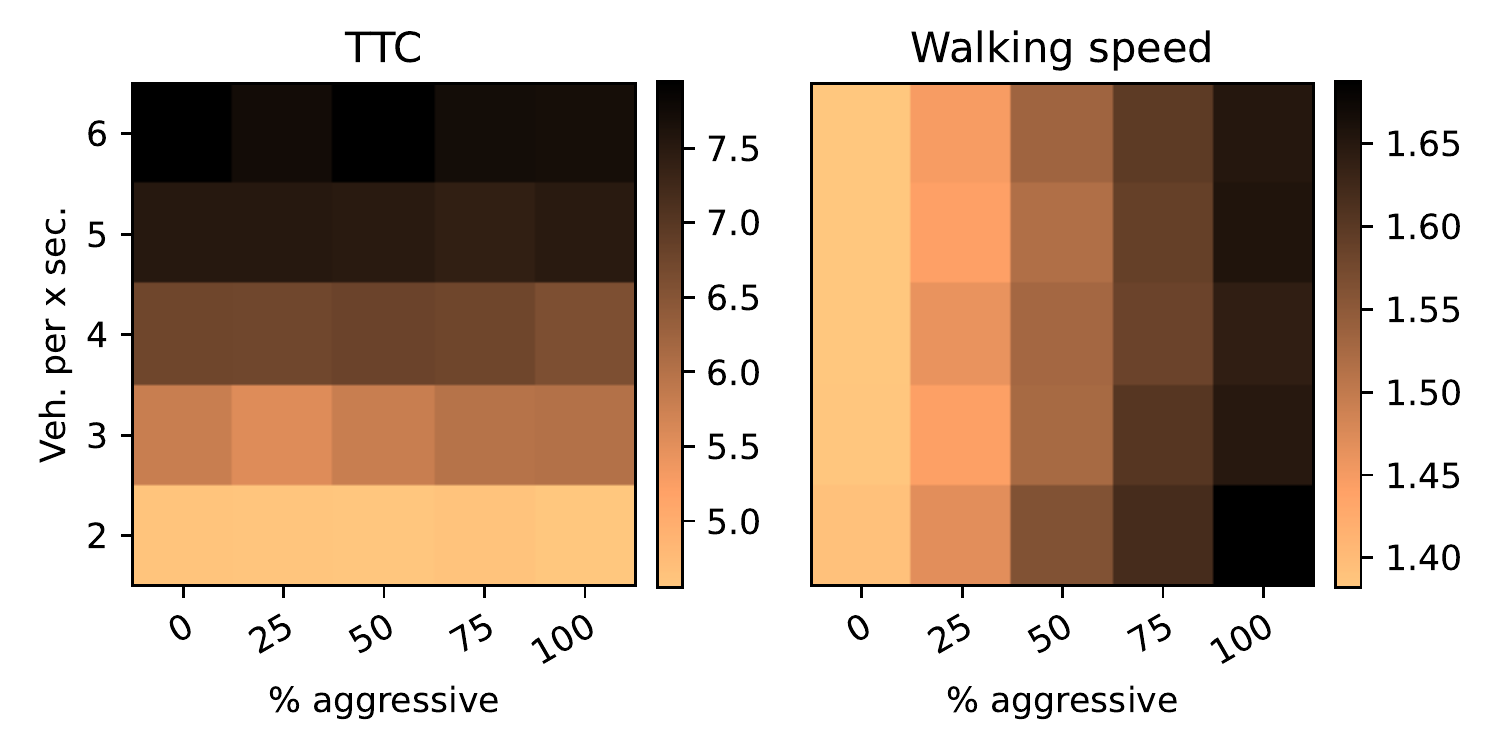}
\caption{}
\label{fig:violating_vs_aggressive}
\end{subfigure}
\caption{a) \% completed illegal crossings for different law obedience and and trait settings (applied to all pedestrians). b) \% of completed illegal crossings by average pedestrians with different proportions of aggressive pedestrians in different traffic conditions. c) mean TTC at the time of crossing (left pane) and mean walking speed (right pane) of violating pedestrians depending on the proportion of aggressive pedestrians in different traffic conditions.}
\vspace{-1em}
\label{fig:traits_vs_lo}
\end{figure}

We start with evaluating the effects of traits and law obedience, the two main characteristics that affect pedestrian decision-making. We ran the simulation for 2500 steps while varying the traits and law-obedience of all pedestrians with the remaining parameters set to default. The road network was fixed to a 4-way intersection with signalized crossings after every other edge. The vehicle generation rate was set to 1 vehicle every 2s to simulate medium-to-heavy traffic flow.

The plot in Figure \ref{fig:lo_vs_trait} shows percentages of completed illegal crossings for each combination of traits and law obedience settings. As expected, setting both to random results in the uniform mix of legal and illegal crossings. Varying law obedience from \textit{violating} to \textit{average} and \textit{obedient} drastically affects jaywalking behaviors. \textit{Average} pedestrians rarely choose to jaywalk ($4-6$\% illegal crossings obtained in simulation matches the statistics in PIE dataset), while \textit{obedient} pedestrians never do so. Traits in combination with law obedience create more behavior patterns. As a result, \textit{aggressive} pedestrians find more jaywalking opportunities owing to their higher walking speed and lower gap acceptance thresholds. 

\begin{table*}[htb!]
\centering
\caption{Effects of TTC method on crossing behavior. Mean and stdev values (in brackets) for wait time, minimum TTC at the time of crossing, and number of vehicle-person collisions are shown for different vehicle generation rates.}
\label{tab:ttc_mode}
\resizebox{\textwidth}{!}{%
\begin{tabular}{l|ccc|ccc|ccc}
Measures & \multicolumn{3}{c|}{Wait time (s)} & \multicolumn{3}{c|}{Min TTC (s)} & \multicolumn{3}{c}{Num. veh-person collisions} \\ \hline
\backslashbox{TTC method}{Traffic} & light & med & heavy & light & med & heavy & light & med & heavy \\ \hline
Constant & 2.7 (3.2) & 2.2 (2.5) & 2.3 (2.1) & 7.5 (4.8) & 6.6 (4.6) & 5.7 (4.7) & 3.2 (2.3) & 5.2 (1.3) & 5.0 (3.9) \\
Average & 5.9 (9.0) & 5.5 (8.3) & 6.5 (9.5) & 5.9 (4.4) & 4.4 (4.1) & 3.9 (3.9) & 4.0 (1.6) & 2.6 (2.3) & 7.0 (2.8) \\ \hline
Dynamic & 3.2 (4.0) & 3.1 (3.6) & 3.5 (4.1) & 7.3 (4.7) & 6.3 (4.7) & 5.2 (4.5) & 3.8 (2.0) & 5.8 (3.1) & 5.2 (2.9) \\
Dynamic+adj & 3.8 (5.2) & 3.6 (3.8) & 4.3 (5.0) & 7.4 (4.6) & 6.6 (4.6) & 5.8 (4.5) & 0.0 (0.0) & 0.4 (0.5) & 0.2 (0.4) \\
Dynamic+adj+noise & 4.3 (5.8) & 4.0 (5.0) & 5.2 (6.2) & 7.8 (4.5) & 7.2 (4.5) & 6.1 (4.5) & 0.4 (0.5) & 0.4 (0.5) & 0.2 (0.4)
\end{tabular}%
}
\vspace{-1em}
\end{table*}

For a fine-grained demonstration of combined effects of traits and law obedience on pedestrian risk-taking and walking behavior, we conducted the following experiment. We fixed the road layout (4-way intersection), TTC computation (dynamic), and varied the vehicle generation rate (1 vehicle every 8, 6, 5, 4, 3s) as well as the percentage of \textit{aggressive} pedestrians from 0\% to 100\% with a step of 25\%. We also changed law obedience for all pedestrians to \textit{average} and \textit{violating}. For every condition, we performed five runs with 1000 steps each (0.1 step duration).

Figure \ref{fig:average_vs_aggressive} shows two trends. First, as the traffic density increases, average pedestrians tend to jaywalk less. Here, pedestrians wait too long for a suitable gap, and more of them tend to cross at the nearest designated crosswalk (see Section \ref{sec:wait_too_long}). Second, as the percentage of aggressive pedestrians increases, pedestrians tend to walk faster and accept shorter gaps, thus leading to more illegal crossings. Similar observations can be made for violating pedestrians: aggressive pedestrians accept smaller gaps (Figure \ref{fig:violating_vs_aggressive} left) as their average walking speed increases (Figure \ref{fig:violating_vs_aggressive} right).

\subsection{TTC computation methods}
TTC estimation plays a large role in pedestrians' decision to cross. Pedestrians who overestimate TTC are at higher risk of accidents, whereas those who underestimate TTC may miss opportunities to cross. For a quantitative evaluation of the effects of different TTC computation methods, we performed the following set of experiments. We fixed the road layout to a 4-way intersection as it is the most complex one for computing the TTC (due to vehicles turning). We set law obedience of all pedestrians to \textit{violating}, their crossing pattern to \textit{one-stage} crossing, and simulated light, medium, and heavy traffic flow by generating 1 vehicle every 6, 4, and 2s, respectively. The remaining parameters were set to default. For every combination of parameters we performed 5 runs with 1000 steps each and computed different versions of the TTC, which include \textit{constant}, \textit{average}, \textit{dynamic}, \textit{dynamic+adj} (with adjustment for pedestrian travel time and the length of the vehicle), and \textit{dynamic+adj+noise} (with adjustment and perceptual noise).

Table \ref{tab:ttc_mode} shows how TTC computation affects pedestrian decision-making represented by the wait time, estimates of minimum TTC at the time of crossing, and, most importantly, the number of collisions between vehicles and pedestrians. 

\textit{Constant} method  consistently overestimates the TTC, leading to the shortest average wait times but also a higher number of collisions. \textit{Average} method tends to underestimate TTC, thus increasing the wait times, but still leads to a significant number of collisions. \textit{Dynamic} method provides the most accurate estimate of TTC, however, pedestrians tend to collide with the sides of the vehicles. Adjusting of TTC calculation (\textit{Dynamic+adj}) nearly eliminates crashes, at a price of a small increase in wait time compared to \textit{Dynamic} method. Few accidents remain mainly due to limitations of the vehicle reactive behavior model. For example, pedestrians crossing very close to intersections may notice vehicles that are about to turn, but the vehicles only see them when their paths are aligned, at which point it is too late to react. Finally, adding perceptual noise results in slightly more conservative TTC estimation, further increasing the wait time, however, the number of accidents remains unchanged. 

Changing traffic conditions creates more behavior patterns. When traffic is heavier, there are fewer gaps available which results in increased wait times. However, as noted in Sections \ref{sec:wait} and \ref{sec:wait_too_long}, pedestrian risk tolerance decreases as they wait. As a result, they tend to accept lower gaps, which is reflected in decreasing min TTC at the time of crossing as the traffic flow density increases. With smaller safety margins, incorrect TTC computation is more likely to lead to accidents. Increased traffic flow also creates more complex interactions between agents, as reflected in higher standard deviations of all metrics. 

\subsection{Crossing patterns}
\begin{figure}[!t]
\centering
\includegraphics[width=1\columnwidth]{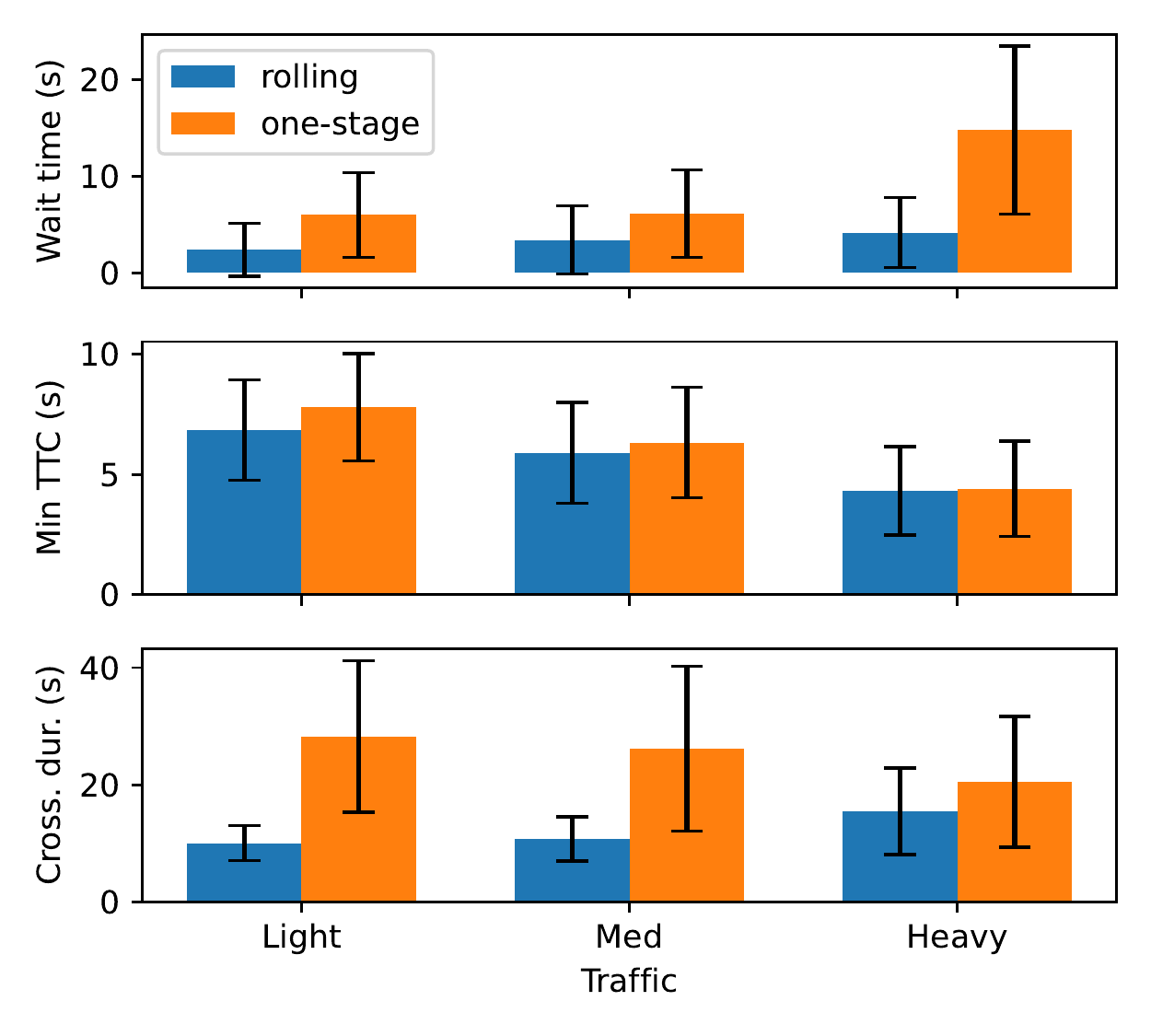}
\caption{Effect of crossing patterns on pedestrians' wait times, TTC at the time of crossing, and duration of crossing for light, medium, and heavy traffic flow.}
\label{fig:crossing_patterns}
\vspace{-1em}
\end{figure}

To test the effects of crossing patterns, we used the same road layout and traffic generation parameters as in the previous section. All pedestrians were set to \textit{violating} and used \textit{dynamic+adj} TTC computation. For each traffic condition, we varied crossing patterns of all pedestrians from \textit{rolling gap} to \textit{one-stage} and measured wait time, TTC at the time of crossing, and duration of the crossing. The results are shown in Figure \ref{fig:crossing_patterns}.

As expected, wait times for \textit{rolling gap} strategy are significantly shorter than those for \textit{one-stage} crossing, since opportunities to start crossing are more frequent if only the nearest lane is taken into account. Furthermore, wait times under \textit{rolling gap} strategy remain approximately the same as in all traffic conditions, whereas pedestrians crossing in \textit{one-stage} see large increases in both the mean and std of waiting duration caused by fewer available gaps as more cars populate the streets.

Likewise, the time it takes pedestrians to cross is significantly longer for \textit{one-stage} strategy for light and medium traffic due to the fact that pedestrians continuously evaluate traffic. If the initial estimate of the gap changes (\eg due to traffic signals or vehicles changing lanes or turning), pedestrians may be stuck in the middle of the road waiting for the suitable gap in the remaining lanes. \textit{Rolling gap} strategy does not suffer from this drawback as pedestrians only look for the opening in a single nearest lane at a time. In heavy traffic, crossing durations are similar for both strategies due to the higher probability of stopped traffic that allows pedestrians to go around blocked vehicles regardless of the crossing strategy.

Finally, TTC at the time of crossing reduces for both crossing strategies as the traffic density increases as a result of smaller gaps between vehicles and longer wait times that increase pedestrians' risk tolerance.

\section{Conclusions and future work}
In this paper, we presented a novel microscopic agent-based model for realistic pedestrian behavior generation. The proposed model includes many factors that affect crossing behavior. These factors are supported by the findings of traffic behavior research and include individual characteristics of pedestrians, the types of choices they make, and how they assess the environment (\eg relevance of other agents and perceptual noise). The parameters of the model are calibrated using both naturalistic driving data and results from the literature. 

Extensive experimental evaluation demonstrates that the proposed model is capable of generating diverse pedestrian behaviors and shows the effects of changing key parameters. These properties serve many application domains, such as intelligent driving systems, transportation planning, and traffic analysis. Our model can be used to study behaviors emerging from interactions of heterogeneous traffic agents and to reconstruct specific scenarios using field data. 

As future work, we plan to extend the vehicle behavior model, implement pedestrian group dynamics, and improve perception models for all agents.

\bibliographystyle{IEEEtran}
\bibliography{references}
\end{document}